\documentclass[10pt,twocolumn,letterpaper]{article}

\usepackage{iccv}
\usepackage{times}
\usepackage{epsfig}
\usepackage{graphicx}
\usepackage{grffile}
\usepackage{amsmath}
\usepackage{amssymb}
\usepackage{booktabs}

\usepackage[pagebackref=true,breaklinks=true,letterpaper=true,colorlinks,bookmarks=false]{hyperref}

\iccvfinalcopy

\ificcvfinal\fi
\usepackage{graphicx}
\usepackage{amsmath,amssymb}  
\usepackage{color}
\usepackage{pgfplots}
\usepackage{tikz}
\usepackage{array}
\usepackage{multirow}
\usepackage{booktabs}
\usepackage{colortbl}
\usepackage{floatrow}
\usepackage{bm}

\DeclareMathOperator*{\argmax}{argmax}
\DeclareMathOperator*{\argmin}{argmin}
\newcommand\myparagraph[1]{\vspace{0pt}\noindent\textbf{#1}\quad}

\def\tablescalingfactor{0.85}

\begin{document}

\title{Neural Activation Constellations: Unsupervised Part Model Discovery with Convolutional Networks}

\author{Marcel Simon and Erik Rodner\\
Computer Vision Group, University of Jena, Germany\thanks{The authors thank NVIDIA for GPU hardware donations.}\\
{\small\url{http://www.inf-cv.uni-jena.de/constellation\_model\_revisited}}}

\maketitle

\begin{abstract}
\vspace{-7pt}
Part models of object categories are essential for challenging recognition tasks,
where differences in categories are subtle and only reflected in appearances of small
parts of the object. We present an approach that is able to learn part models in a 
completely unsupervised manner, without part annotations and even without given bounding boxes
during learning. The key idea is to find constellations of neural activation patterns
computed using convolutional neural networks. 
In our experiments, we outperform existing approaches for fine-grained
recognition on the CUB200-2011, NA birds, Oxford PETS, and Oxford Flowers dataset in case no part or bounding box annotations 
are available and achieve state-of-the-art performance for the Stanford Dog dataset. 
We also show the benefits of neural constellation models as a data augmentation technique for fine-tuning.
Furthermore, our paper unites the areas of generic and fine-grained classification, since our approach
is suitable for both scenarios. 
\end{abstract}

\vspace{-0pt}
\section{Introduction}
\vspace{-5pt}
Object parts play a crucial role in many recent approaches for fine-grained recognition.
They allow for capturing very localized discriminative features of an object~\cite{Goering14:NPT}.
Learning part models is often either done in a completely supervised manner by providing part annotations~\cite{branson14cub75acc,zhang12-ppk}
or labeled bounding boxes~\cite{dpm,Simon14:PDD}. 

\begin{figure}[t]
 \centering
  \includegraphics[width=0.7\linewidth]{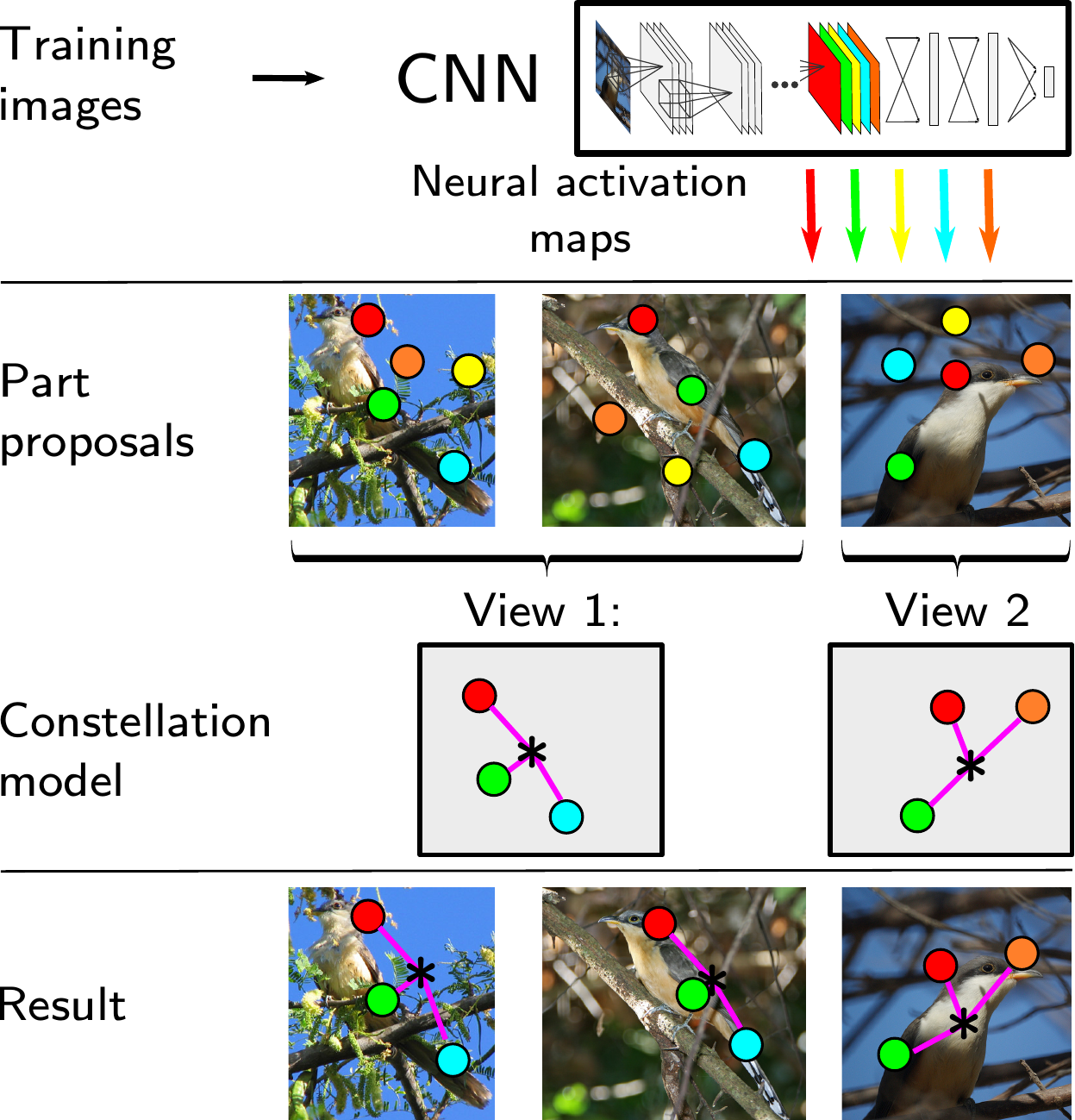}
 \caption{Overview of our approach. 
 Deep neural activation maps are used to exploit the channels of a CNN as a part detector. 
 We estimate a part model from completely unsupervised data by selecting part detectors that fire at similar relative locations. 
 The created part models are then used to extract features at object parts for weakly-supervised classification.
 \label{fig:teaser}
 }
\end{figure}

In contrast, we show how to learn part-models in a completely unsupervised manner, which drastically reduces annotation costs for learning.
Our approach is based on learning constellations of neural activation patterns obtained from pre-learned convolutional neural networks (CNN).
Fig.~\ref{fig:teaser} shows an overview of our approach. Our part hypotheses are outputs of an intermediate CNN layer for
which we compute neural activation maps~\cite{Simon14:PDD,SimonyanSaliency}. Unsupervised part models are either build by
randomly selecting a subset of the part hypotheses or learned by estimating the parameters of a generative spatial part
model. In the latter case, we implicitly find subsets of part hypotheses that ``fire'' consistently in a certain constellation in the images.

Although creating a model for the spatial relationship of parts has already been introduced a decade ago~\cite{fergus2003classRecog,fei2006oneShot},
these approaches face major difficulties due to the fact that part proposals
are based on hand-engineered local descriptors and detectors without correspondence
We overcome this problem by using implicit part detectors of a pre-learned CNN, which at the same time greatly simplifies the part-model training.
As shown by \cite{zeiler2013visualizing}, intermediate CNN outputs can often be linked to semantic parts of common objects and
we are therefore using them as part proposals.
Our part model learning has to select only a few parts for each view of an object from an already high quality pool of part proposals.
This allows for a much simpler and faster part model creation without the need to explicitly consider appearance of the individual parts as done in previous works~\cite{fergus2003classRecog,andriluka}.
At the same time, we do not need any ground-truth part locations or bounding boxes.

The obtained approach and learning algorithm improves the state-of-the-art in fine-grained recognition on three datasets including CUB200-2011~\cite{wahCUB_200_2011} if no ground-truth part or bounding box annotations are available at all.
In addition, we show how to use the same approach for generic object recognition on Caltech-256.
This is a major difference to previous work on fine-grained recognition, since most approaches are not directly applicable to other tasks. 
For example, our approach is able to achieve state-of-the-art performance on Caltech-256 without the need for expensive dense evaluation on different scales of the image~\cite{Simonyan14veryDeep}. 

Furthermore, our work has impact beyond fine-grained recognition, since our method can also be used to guide data augmentation during fine-tuning for image classification.
We demonstrate in our experiments that it even yields a more discriminative CNN compared to a CNN fine-tuned with ground-truth bounding boxes of the object.

In the next section, we give a brief overview over recent approaches in the areas of part constellation models and fine-grained classification.
Sect.~\ref{sec:neural_maps} reviews the approach of Simon~\etal~\cite{Simon14:PDD} for part proposal generation.
In Sect.~\ref{sec:part_discovery}, we present our flexible unsupervised part discovery method. 
The remaining paper is dedicated to the experiments on several datasets (Sect.~\ref{sec:experiments}) and conclusions (Sect.~\ref{sec:conclusions}).

\section{Related work\label{sec:related_work}}
\vspace{-5pt}
    \myparagraph{Part constellation models}
      Part constellation models describe the spatial relationship between object parts.
      There are many supervised methods for part model learning which rely on ground-truth part or bounding box annotations~\cite{zhang2013dpd,Goering14:NPT,Simon14:PDD}.
      However, annotations are often not available or expensive to obtain. 
      In contrast, the unsupervised setting does not require any annotation and relies on part proposals instead.
      It greatly differs from the supervised setting as the selection of useful parts is crucial.
      We focus on unsupervised approaches as these are the most related to our work.

      One of the early works in this area is \cite{zobel2000robust}, where facial landmark detection was done by fusing single detections with a coupled ray model.     
      Similar to our approach, a common reference point is used and the position of the other parts are described by a distribution of their relative polar coordinates. 
      However, they rely on manually annotated parts while we focus on the unsupervised setting.
      Later on, Fergus~\etal~\cite{fergus2003classRecog} and Fei-Fei~\etal~\cite{fei2006oneShot} build models based on generic SIFT interest point detections. 
      The model includes the relative positions of the object parts as well as their relative scale and appearance.
      While their interest point detector delivers a number of detections without any semantics, each of the CNN-based part detectors we use correspond to a specific object part proposal already.
      This allows us to design the part selection much more efficient and to speed up the inference.
      The run time complexity compared to \cite{fergus2003classRecog,fei2006oneShot} decreases from exponential in the number of modeled parts to linear time complexity.
      Similar computational limitations occur in other works as well, for example \cite{riabchenko14constellation}.
      Especially in the case of a large number of part proposals this is a significant benefit. 
      
      Yang~\etal~\cite{templatenips} select object part templates from a set of randomly initialized image patches.
      They build a part model based on co-occurrence, diversity, and fitness of the templates in a set of training images. 
      The detected object parts are used for part-based fine-grained classification of birds.
      In our application, co-occurrence and fitness are rather weak properties for the selection of CNN-based part 
      proposals.
      For example, detectors of frequently occurring background patterns such as leaves of a tree would likely 
      be selected by their algorithm. 
      Instead our work considers the spatial relationship in order to filter unrelated background detectors 
      that fire on inconsistent relative locations. 
      
      Crandall~\etal~\cite{crandall07composite} improve part model learning by jointly considering object and scene-related parts.
      However, the number of combinations of possible views of an object and different background patterns is huge.
      In contrast, our approach selects the part proposals based on the relative positions which is simpler and effective since we only want to identify useful part proposals for classification.
      
      In the area of detection, there are numerous approaches based on object parts.
      The deformable part model (DPM, \cite{dpm}) is the most popular one.
      It learns part constellation models relative to the bounding box with a latent discriminative SVM model.
      Most detection methods require at least ground-truth bounding box annotations. 
      In contrast, our approach does not require such annotations or any negative examples, since we learn
      the constellation model in a generative manner and by using object part proposals not restricted to a bounding box.

    \myparagraph{Fine-grained recognition with part models}
	Fine-grained recognition focuses on visually very similar classes,
    where the different object categories sometimes differ only in minor details.
	Examples are bird species~\cite{wahCUB_200_2011} or car models~\cite{krause2013cars} recognition.
	Since the differences of small parts of the objects matter, localized feature extraction using
    a part model plays an important role.

	One of the earliest work in the area of fine-grained recognition uses an ellipsoid to model the bird pose~\cite{farrell11-bsc} and fuse obtained parts using very specific kernel functions~\cite{zhang12-ppk}. 
	Other works build on deformable part models~\cite{dpm}. For example,
	the deformable part descriptor method of \cite{zhang2013dpd} uses a supervised version of \cite{dpm}
    for training deformable part models, which then allows for pose normalization by comparing corresponding parts.
	The work of \cite{gavves2013alignment} and \cite{Goering14:NPT} demonstrated nonparametric part detection for fine-grained recognition. 
	The basic idea is to transfer human-annotated part positions from similar training examples obtained with nearest neighbor matching. 
	Chai~\etal~\cite{chai2013symbiotic} use the detections of DPM and the segmentation output of GrabCut to predict part locations.
	Branson~\etal~\cite{branson14cub75acc} use the part locations to warp image patches into a pose-normalized representation.
	Zhang~\etal~\cite{zhang2014partRCNN} select object part detections from object proposals generated by Selective Search~\cite{uijlings13selSeach}.
	The mentioned methods use the obtained part locations to calculate localized features.
	Berg~\etal~\cite{berg2013partclass} learns a linear classifier for each pair of parts and classes. 
	The decision values from numerous of such classifiers are used as feature representation.
	While all these approaches work well in many tasks, they require ground-truth part annotations at training and often also at test time.
	In contrast, our approach does not rely on expensive annotated part locations and is fully unsupervised for part model learning instead.
	This also follows the recent shift of interest towards less annotation during training \cite{zhang2014partRCNN,xiao14attention,Simon14:PDD}.
	The method of Simon~\etal~\cite{Simon14:PDD} presents a method, which
	requires bounding boxes of the object during training rather than part annotations.
	They also make use of neural activation maps for part discovery, but although our approach does not
	need bounding boxes we are still able to improve over their results. 

	The unsupervised scenario that we tackle has also been considered by Xiao~\etal~\cite{xiao14attention}.
	They cluster the channels of the last convolutional layers of a CNN into groups. 
	Patches for the object and each part are extracted based on the activation of each of these groups.
	The patches are used to classify the image.
	While their work requires a pre-trained classifier for the objects of interest, 
    we only need a CNN that can be pre-trained on a weakly related object dataset.

\section{Deep neural activation maps\label{sec:neural_maps}}
\vspace{-5pt}
CNNs have demonstrated an amazing potential to learn a complete classification pipeline from scratch without the need to manually define low level features.
Recent CNN architectures \cite{krizhevsky2012imagenet,Simonyan14veryDeep} consist of multiple layers of convolutions, pooling operations, 
full linear transformations and non-linear activations.

The convolutional layers convolve the input with numerous kernels. 
As shown by \cite{zeiler2013visualizing}, the kernels of the convolutions in early layers are similar to the filter masks used in many popular 
low level feature descriptors like HOG or SIFT.
Their work also shows that the later layers are sensitive to increasingly abstract patterns in the image.
These patterns can even correspond to whole objects~\cite{SimonyanSaliency} or parts of objects~\cite{Simon14:PDD} and this is exactly what we exploit.

\begin{figure}
\centering
  {\scriptsize
  \setlength{\tabcolsep}{2pt}
  \setlength{\arraycolsep}{2pt}
  \begin{tabular}{ccc}
    \includegraphics[width=0.25\linewidth]{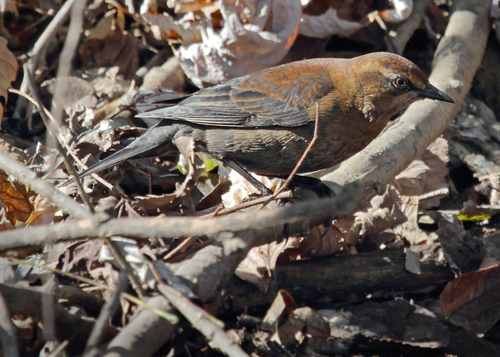} &
    \includegraphics[width=0.25\linewidth]{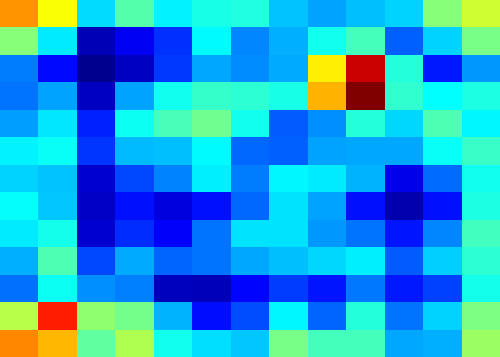} &
    \includegraphics[width=0.25\linewidth]{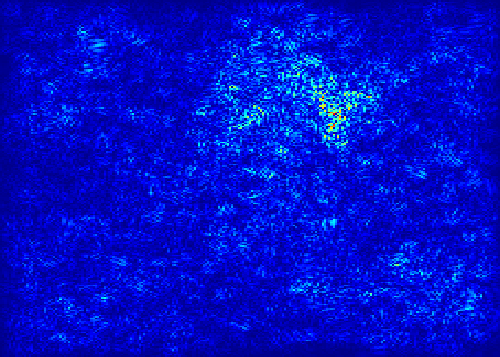} \\
    \includegraphics[width=0.25\linewidth]{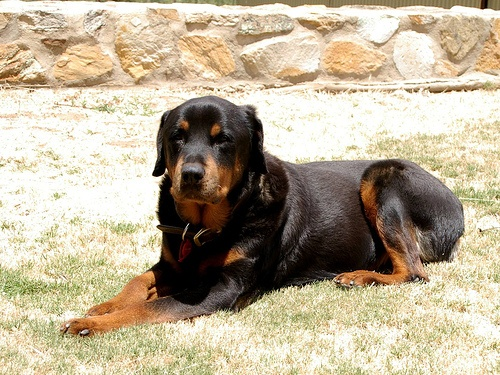} &
    \includegraphics[width=0.25\linewidth]{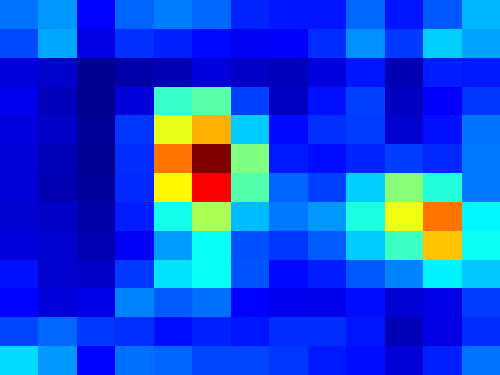} &
    \includegraphics[width=0.25\linewidth]{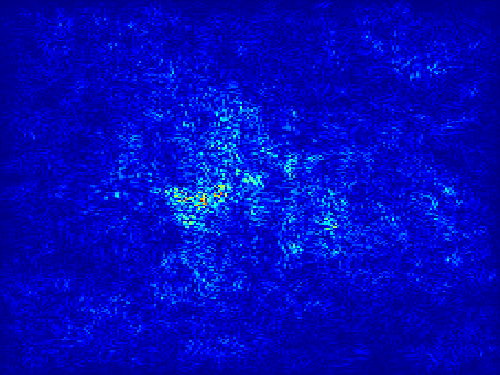}\\
    &&\\
    Input & CNN last conv. output & Neural activation map\\
    $\bm{I}$ & 
    \resizebox{0.32\linewidth}{0.06\linewidth}{$\left[ \begin{array}{ccc}  f_{1,1} & \ldots & f_{1,13}\\ \ldots & \ldots & \ldots\\ f_{13,1} & \ldots & f_{13,13} \end{array} \right]$ }&
    \resizebox{0.32\linewidth}{0.07\linewidth}{$\left[ \begin{array}{ccc}  m_{1,1} & \ldots & m_{1,227}\\ \ldots & \ldots & \ldots\\ m_{227,1} & \ldots & m_{227,227} \end{array} \right]$}
  \end{tabular}
  \setlength{\tabcolsep}{8pt}
  }
\caption{Examples for the output of a channel of the last convolutional layer and the corresponding neural activation maps for two images (index of the channel is skipped to ease
notation). A deep red corresponds to high activation and a deep blue to no activation at all. Activation maps are available in higher resolution and better suited 
for part localization. Best viewed in color.
\label{fig:pool5_outputs}}
\end{figure}

The output $f$ of a layer before the fully-connected layers is organized in multiple channels $1 \leq p \leq P$ with a two-dimensional arrangement of output elements,
\ie we denote $f$ by $( f_{j,j'}^{(p)}(\bm{I}) )$ where $\bm{I} \in \mathbb{R}^{W \times H}$ denotes the input image and $j$ and $j'$ are indices of the output elements in the channel. 
Fig.~\ref{fig:pool5_outputs} shows examples of such a channel output for the last convolutional layer.
As can be seen the output can be interpreted as detection scores of multiple object part detectors.
Therefore, the CNN automatically learned implicit part detectors relevant for the dataset it was trained from. 
In this case, the visualized channel shows high outputs at locations corresponding to the head of birds and dogs.

A disadvantage of the channel output is its resolution, which would not allow for precise localization of parts.
Due to this reason, we follow the basic idea of \cite{SimonyanSaliency} and \cite{Simon14:PDD} and
compute \emph{deep neural activation maps}. We calculate the gradient
of the average output of the channel $p$ with respect to the input image pixels $I_{x,y}$:
\vspace{-5pt}
\begin{align}
m_{x,y}^{(p)}(\bm{I}) &= \frac{\partial}{\partial I_{x,y}} \sum\limits_{j,j'} f_{j,j'}^{(p)}(\bm{I}) 
\end{align}
\vspace{-2pt}
The calculation can be easily achieved with a back-propagation pass~\cite{Simon14:PDD}.
The absolute value of the gradient shows which pixels in the image have the largest impact on the
output of the channel.  Similar to the actual output of the layer, it allows for localizing image areas this channel is sensitive to.
However, the resolution of the deep neural activation maps is much higher (Fig.~\ref{fig:pool5_outputs}). 
In our experiments, we compute part proposal locations for a training image $\bm{I}_{i}$ from these maps by 
using the point of maximum activation:
\begin{align}
\bm{\mu}_{i,p} &= \argmax_{x,y} \left| m^{(p)}_{x,y}(\bm{I}_{i}) \right|.
\end{align}
Each channel of the CNN delivers one neural activation map per image and we therefore obtain one part proposal per channel $p$.
RGB images are handled by adding the absolute activation maps of each input channel.
Hence we reduce a deep neural activation map to a 2D location
and do not consider image patches for each part during the part model learning.
In classification, however, image patches are extracted at predicted part locations for feature extraction.

The implicit part detectors are learned automatically during the training of the CNN.
This is a huge benefit compared to other part discovery approaches like poselets~\cite{poselets}, which do not necessarily produce parts useful for discrimination of classes a priori.
In our case, the dataset used to train the CNN does not necessarily need to be the same as the final dataset
and task for which we want to build part representations. 
In addition, determining the part proposals is nearly as fast as the classification with the CNN (only $110$ms per image for $10$ parts on a standard PC with GPU),
which allows for real-time applications.
A video visualizing a bird head detector based on this idea running at 10fps is available at our project website.
We use the part proposals throughout the rest of this paper.

\vspace{-5pt}
\section{Unsupervised part model discovery\label{sec:part_discovery}}
\vspace{-5pt}
In this section, we show how to construct effective part models in an unsupervised manner 
given a set of training images of an object class.
The resulting part model is used for localized feature extraction and subsequent fine-grained classification.
In contrast to most previous work, we have a set of robust but not necessarily related part proposals and need to select useful ones for the current object class. 
Other approaches like DPM are faced with learning part detectors instead.
The main consequence is that we do not need to care about expensive training of robust part detectors.
Our task simplifies to a selection of useful detectors instead. 

As input, we use the normalized part proposal locations $\bm{\mu}_{i,p}\in\left[0,1\right]^{2}$
for training image $i=1,\dots,N$ and part proposal $p=1,\dots,P$. The $P$ part proposals
correspond to the channels an intermediate output layer in a CNN and $\bm{\mu}_{i,p}$ is determined
by calculating the activation map of channel $p$ for input image
$i$ and locating the maximum response. If the activation map of a channel is equal to $\bm{0}$, the part proposal
is considered hidden. This sparsity naturally occurs due to the rectified linear unit used as a nonlinear activation.

\subsection{Random selection of parts\label{sec:random_selection}}
A simple method to build a part model with multiple parts is to select $M$ random parts from all $P$ proposals.
For all training images, we then extract $M$ feature vectors describing the image region around the
part location. The features are stacked and
a linear SVM is learned using image labels. This can even be combined with fine-tuning of the CNN
used to extract the part features. Further details about part feature representations are given in Sect.~\ref{sec:experiments}.

In our experiments, we show that for generic object recognition random selection is indeed a valid technique. However,
for fine-grained recognition, we need to select the parts that likely correspond to the same object and not a background artifact.
Furthermore, using all proposals is not an option since the feature representation increases dramatically rendering training impractical. Therefore, we show in the following how to select only a few parts with a constellation model to boost classification
performance and reduce computation time for feature calculation significantly.

\subsection{Constellations of neural activations\label{sec:part_constellation}}
The goal is to estimate a star shape model for a subset of selected proposals
using the 2D locations of all part proposals of all training images.
Similar to other popular part models like DPM~\cite{dpm}, our model also incorporates multiple views $v=1,\dots,V$
of the object of interest. For example, the front and the side view
of a car is different and different parts are required to
describe each view. 

Each view consists of a selection of $M$ part proposals denoted by the indicator
variables $b_{v,p} \in \left\{ 0,1\right\}$ and we refer to them as parts.
In addition, there is a set of corresponding shift vectors $\bm{d}_{v,p}\in\left[-1,1\right]^{2}$.
The shift vectors are the ideal relative offset of part $p$ to the common root
location $\bm{a}_{i}$ of the object in image $i$.
The $\bm{a}_i$ are latent variables since no object annotations are given during learning.

Another set of latent variables $s_{i,v}\in\left\{ 0,1\right\}$
denotes the view selection for each training
image. We assume that there is only one target object visible in each image
and hence only one view is selected for each image. Finally, $h_{i,p}\in\left\{ 0,1\right\}$
denotes if part $p$ is visible in image
$i$. In our case, the visibility of a part is provided by the part
proposals and not estimated during learning.

\myparagraph{Learning objective}
We identify the best model for the given training images
by maximum a-posteriori estimation of all model and latent parameters 
$\bm{\Gamma} = (\bm{b}, \bm{d}, \bm{s}, \bm{a})$ from provided part proposal locations $\bm{\mu}$:
\vspace{-3pt}
\begin{equation}
\bm{\hat{\Gamma}} = \textstyle\argmax_{ \bm{\Gamma} } \quad
p\left(\bm{\Gamma} \;|\; \bm{\mu}\right).\label{eq:optimization-1}
\end{equation}
\vspace{-2pt}
In contrast to a marginalization of the latent variables, we obtain a very efficient learning algorithm.
We apply Bayes' rule, use the typical assumption that training images and part proposals are independent 
given the model parameters~\cite{andriluka}, assume flat priors for $\bm{a}$ (no prior preference for the object's center)
and $\bm{d}$ (no prior preference for part offsets), and independent priors for $\bm{b}$ and $\bm{s}$:
\vspace{-3pt}
\begin{align}
 & \argmax_{\bm{\Gamma}}\; p\left(\bm{\mu} \,|\, \bm{b}, \bm{d}, \bm{s}, \bm{a}\right)\cdot p(\bm{b}) \cdot p(\bm{s}) \nonumber \\
= & \argmax_{\bm{\Gamma}}\; \prod_{i=1}^{N} \left( \prod_{p=1}^{P} p\left(\bm{\mu}_{i,p} \,|\, \bm{b}, \bm{d}, \bm{s}, \bm{a}\right) \right) p(\bm{b}) \cdot p(\bm{s})\label{eq:optimization-2}
\end{align}
\vspace{-2pt}
The term $p\left(\bm{\mu}_{i,p} \,|\, \bm{b},\bm{d}, \bm{s}, \bm{a}\right)$ is
the distribution of the predicted part locations given the model.
If the part $p$ is used in view $v$ of image $i$, we assume
that the part location is normally distribution around the root location plus the shift vector, \ie $\bm{\mu}_{i,p} \sim \mathcal{N}(\bm{d}_{v,p} + \bm{a}_i, \sigma^2_{v,p} \bm{E})$
with $\bm{E}$ denoting the identity matrix. 
If the part is not used, there is no
prior information about the location and we assume it to be uniformly
distributed over all possible image locations in $\bm{I}_i$. Hence, the distribution is given by
\begin{align}
&p\left(\bm{\mu}_{i,p} \,|\, \bm{b}, \bm{d}, \bm{s}, \bm{a}\right) =\\
\notag
&\prod\limits_{v=1}^V \mathcal{N}\left( \bm{\mu}_{i,p} \,|\, \bm{a}_{i}+\bm{d}_{v,p}, \sigma^2_{v,p} \bm{E}\right)^{t_{i,v,p}} \cdot \left(\frac{1}{\left|\bm{I}_{i}\right|}\right)^{1-t_{i,v,p}},
\end{align}
 where $t_{i,v,p}=s_{i,v} b_{v,p} h_{i,p} \in\left\{ 0,1\right\} $
indicates whether part $p$ is used and visible in view $v$ which is itself active in image $i$. 
The prior distribution for the part selection $\bm{b}$ only 
captures the constraint that $M$ parts need to be selected, \ie $\forall v: M = \sum_{p=1}^P b_{v,p}$.
The prior for the view selection $\bm{s}$ incorporates our assumption that only a single view
is active in training image $i$, \ie $\forall i: 1 = \sum_{v=1}^V s_{i,v}$. In general, 
we denote the feasible set of variables as $\mathcal{M}$.
Exploiting this and applying $\log$ simplifies Eq.~\eqref{eq:optimization-2}
further: 
\begin{align}
\notag
&\argmin_{\bm{\Gamma} \in \mathcal{M}} -\sum_{i=1}^{N}\sum_{p=1}^{P} \sum_{v=1}^V t_{i,v,p}\log
\mathcal{N}\left( \bm{\mu}_{i,p} \,|\, \bm{a}_{i}+\bm{d}_{v,p}, \sigma^2_{v,p}\right)
\end{align}
In addition, we assume the variance
$\sigma_{v,p}^2$ to be constant for all parts of all views. Hence, the final formulation of the optimization problem
becomes 
\begin{align}
& \argmin_{\bm{\Gamma} \in \mathcal{M}} \sum_{i=1}^{N}\sum_{p=1}^{P}\sum_{v=1}^{V}s_{i,v}b_{v,p}h_{i,p}\left\Vert \bm{\mu}_{i,p}-\bm{a}_{i}-\bm{d}_{v,p}\right\Vert ^{2}\label{eq:opt_problem}
\end{align}

\myparagraph{Optimization}
Eq.~\eqref{eq:opt_problem} is solved by alternately optimizing each
of the model variables $\bm{b}$ and $\bm{d}$, as well as the latent variables $\bm{a}$ and $\bm{s}$,
independently, similar to the standard EM algorithm. For each of the variables $\bm{b}$ and $\bm{s}$, we can calculate
the optimal value by sorting error terms. For example, $b_{v,p}$ is
calculated by analyzing 
\vspace{-2pt}
\begin{equation}
\hspace{-8pt}\argmin_{\bm{b} \in \Gamma_{\bm{b}}} \sum_{p=1}^{P}\sum_{v=1}^{V} b_{v,p} \underbrace{\Bigl(\sum_{i=1}^{N} s_{i,v} h_{i,p} \left\Vert \bm{\mu}_{i}^{p}-a_{i}-\bm{d}_{v,p}\right\Vert ^{2} \Bigr)}_{E(v,p)}
\label{eq:optimization_b}
\end{equation}
\vspace{-2pt}
This optimization can be intuitively solved. First, each view is considered independently, as we select a fixed number of parts
for each view without considering the others. For each part proposal, we calculate
$E\left(v,p\right)$. This term describes, how well the part proposal $p$
fits to the view $v$. If its value is small, then the part proposal fits well
to the view and should be selected. We now calculate $E\left(v,p\right)$
for all parts of view $v$ and select the $M$ parts with the smallest
value. In a similar manner, the view selection $\bm{s}$ can be determined. 

The root points $\bm{a}$ are obtained for fixed $\bm{b}, \bm{s}$, and $\bm{d}$ by
\begin{equation}
\hat{\bm{a}}_{i} = \sum_{v,p} t_{i,v,p} \left( \bm{\mu}_{i}^{p}-\bm{d}_{v,p}\right) / \bigl( \sum\limits_{v',p'} t_{i,v',p'} \bigr).
\label{eq:optimization_a}
\end{equation}
Similarly, we obtain the shift vectors $\hat{\bm{d}}_{v,p}$:
\vspace{-2pt}
\begin{equation}
\hat{\bm{d}}_{v,p}= \sum_{i=1}^N t_{i',v,p}\cdot \left( \bm{\mu}_{i,p}-\bm{a}_{i}\right) / \bigl( \sum\limits_{i'=1}^N t_{i',v,p} \bigr).
\end{equation}
\vspace{-2pt}
 The formulas are intuitive as, for example, the shift vectors $\bm{d}_{v,p}$
are assigned the mean offset between root point $\bm{a}_{i}$ and predicted
part location $\bm{\mu}_{i,p}$. The mean, however, is only calculated
for images in which part $p$ is used.

This kind of optimization is comparable to the EM-algorithm and thus shares the same challenges. 
Especially the initialization of the variables is crucial.
We initialize $\bm{a}$ to be the center of the image and $\bm{s}$ as well as $\bm{b}$ randomly to an
assignment of views and selection of parts for each view, respectively. 
The initialization of $\bm{d}$ is avoided by calculating it first.
The value of $\bm{b}$ is used to determine convergence. 
This optimization is repeated with different initializations and the result with the best objective value is used.

\myparagraph{Inference}
The inference step for an unseen test image is similar to the calculations during training. 
The parameters $\bm{s}$ and $\bm{a}$ are iteratively estimated
by solving Eq.~(\ref{eq:optimization_b}) and (\ref{eq:optimization_a})
for fixed learned model parameters $\bm{b}$ and $\bm{d}$. The visibility is again
provided directly by the neural activation maps.

\section{Experiments\label{sec:experiments}}
\vspace{-5pt}
The experiments cover three main aspects and applications of our approach. 
First, we present a data augmentation technique based on the part models of our approach for fine-tuning, which outperforms fine-tuning on bounding boxes.
Second, we apply our approach to fine-grained classification, a task in which most current approaches rely on ground-truth part annotations~\cite{branson14cub75acc,zhang2014partRCNN,Simon14:PDD}.
Finally, we show how to use the same approach for generic image classification, too, and present the benefits in this area. Code for our method will be made available.

\subsection{Experimental setup}
\vspace{-5pt}
\myparagraph{Datasets}
We use five different datasets in the experiments. 
For fine-grained classification, we evaluate our approach on CUB200-2011 \cite{wahCUB_200_2011} (200 classes, 11788 images), NA birds~\cite{horn15nabirds} (555 classes, 48562 images), Stanford dogs~\cite{khosla11stanfordDogs} (120 classes, 20580 images), Oxford flowers 102~\cite{nilsback08-afc} (102 classes, 8189 images), and Oxford-IIIT Pets~\cite{parkhi12-cd} (37 classes, 7349 images).
We use the provided split into training and test and follow the evaluation protocol of the corresponding papers.
Hence we report the overall accuracy on CUB200-2001 and the mean class-wise accuracy on all other datasets.
For the task of generic object recognition, we evaluate on Caltech 256 \cite{cal256scenesurl}, which contains 30607 images of a diverse set of 256 common objects.
We follow the evaluation protocol of \cite{Simonyan14veryDeep} and randomly select 60 training images and use the rest for testing.

\myparagraph{CNNs and parameters}
Two different CNN architectures were used in our experiments: the widely used architecture of Krizhevsky~\etal~\cite{krizhevsky2012imagenet} (AlexNet) and the more accurate one of Simonyan~\etal~\cite{Simonyan14veryDeep} (VGG19).
In case of NA birds, we use GoogLeNet~\cite{szegedy2014going}.
For details about the architecture, we kindly refer the reader to the corresponding papers.
It is important to note that our approach can be used with any CNN.
Features were calculated using the \emph{relu6}, \emph{relu7} and \emph{pool5/7x7\_s1} layer, respectively.
For the localization of parts, the \emph{pool5} layer was used. 
This layer consists of 256 and 512 channels resulting in 256 and 512 part proposals, respectively.
In case of the CUB200-2011, NA birds, Oxford dogs, pets and flowers datasets, fine-tuning with our proposed data augmentation technique is used.   
We use two-step fine-tuning \cite{branson14cub75acc} starting with a learning rate of $0.001$ and decrease it to $0.0001$ when there is no change in the loss anymore.
In case of Stanford dogs, the evaluation with CNNs pre-trained on ILSVRC 2012 images is biased as the complete dataset is a subset of the ILSVRC 2012 training image set. 
Hence, we remove the testing images of Stanford dogs from the training set of ILSVRC 2012 and learned a CNN from scratch on this modified dataset. 
The trained model is available on our website for easy comparison with this work.

If not mentioned otherwise, the learned part models use 5 views and 10 parts per view.
A model is learned for each class separately.
The part model learning is repeated 5 times and the model with the best objective value was taken.
We count in how many images each part is used and select the 10 most often selected parts for use in classification.

\myparagraph{Classification framework}
We use the part-based classification approach presented by Simon~\etal~\cite{Simon14:PDD}.
Given the predicted localization of all selected parts, we crop square boxes centered at each part and calculate features for all of them. 
The size of these boxes is given by $\sqrt{\lambda \cdot W \cdot H}$, $\lambda \in \left\{\frac{1}{5},\frac{1}{16}\right\}$, where $W$ and $H$ are the width and height of the uncropped image, respectively.
If a part is not visible, the features calculated on a mean image are used instead. 
This kind of imputation has comparable performance to zero imputation, but yields in a slight performance gain in some cases. 
In case of CUB200-2011, we also estimate a bounding box for each image.
Selective Search~\cite{uijlings13selSeach} is applied to each image to generate bounding box proposals.
Each proposal is classified by the CNN and the proposal with the highest classification confidence is used as estimated bounding box.

The features of each part, the uncropped image and the estimated bounding box are stacked and classified using a linear SVM.
In case of CUB200-2011, flipped training images were used as well.
Hyperparameters were optimized using cross-validation on the training data of CUB200-2011 and used for the other datasets as well.

\subsection{Data augmentation using part proposals}
\vspace{-5pt}
Fine-tuning is the adaption of a pre-learned CNN to a domain specific dataset.
It significantly boosts the performance in many tasks~\cite{azizpour14CNNanalysis}.
Since the domain specific datasets are often small and thus the training of a CNN is prone to overfitting, the training set is artificially enlarged by using ``data augmentation''.
A common technique used for example by \cite{krizhevsky2012imagenet,Simonyan14veryDeep} is random cropping of a large fixed sized image patch.
This is especially effective if the training images are cropped to the object of interest.
If the images are not cropped and no ground-truth bounding box is available, uncropped images can be used instead.
However, fine-tuning is less effective as shown in Tab.~\ref{tab:augmentation_performance}.
Since ground-truth bounding box annotations are often not available or expensive to obtain, we propose to fine-tune on object proposals filtered by a novel selection scheme instead. 

\begin{figure}
 \centering
 \includegraphics[width=0.7\linewidth]{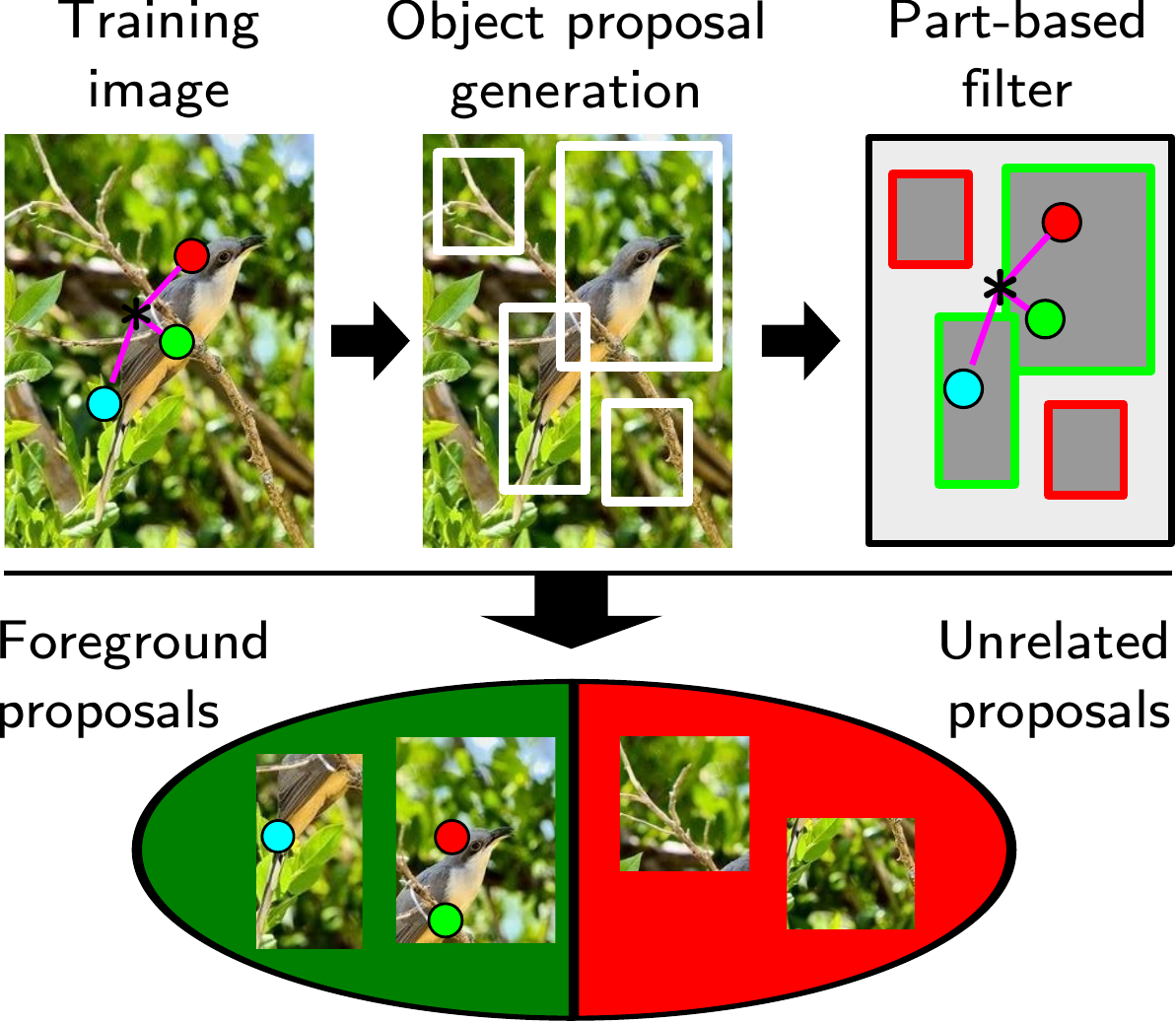}
 \caption{Overview of our approach to filter object proposals for fine-tuning of CNNs.
 Best viewed in color.
 \label{fig:patch_filter}}
\end{figure}

An overview of our approach is shown in Fig.~\ref{fig:patch_filter}.

First, we select for each training image the five parts of the corresponding view, which fit the model best.
Second, numerous object proposals are generated using Selective Search~\cite{uijlings13selSeach}. 
These proposals are very noisy, \ie many only contain background and not the object of interest. 
We count how many of the predicted parts are inside of each proposal and select only proposals containing at least three parts.
The remaining patches, $\approx 48$ on average in case of CUB200-2011, are high quality image regions containing the object of interest. 
Finally, fine-tuning is performed using the filtered proposals of all training images.

\begin{table}
  \centering
\scalebox{\tablescalingfactor}{
  \begin{tabular}{llc}
  \toprule
  Train. Anno.     &Method & Accuracy\tabularnewline
  \midrule
  Bbox&Fine-tuning on cropped images& 67.24\%\tabularnewline
  \midrule
  None&No fine-tuning & 63.77\%\tabularnewline
  None&Fine-tuning on uncropped images& 66.10\%\tabularnewline
  \textbf{None}&Fine-tuning on filtered part proposals& \textbf{67.97\%}\tabularnewline
  \bottomrule
  \end{tabular}
  }
  \caption{Influence of the augmentation technique used for fine-tuning in case of AlexNet on CUB200-2011. 
  Classification accuracies were obtained by using 8 parts as described in Sect.~\ref{sec:experiments_finegrained}.
  \label{tab:augmentation_performance}}
\end{table}
The result of this approach is shown in Tab.~\ref{tab:augmentation_performance}.
Fine-tuning on these patches provides not only a gain even compared to fine-tuning on cropped images, it also eliminates the need for ground-truth bonding box annotations.

\subsection{Fine-grained recognition without annotations\label{sec:experiments_finegrained}}
\vspace{-5pt}
Most approaches in the area of fine-grained recognition rely on additional annotation like ground-truth part locations or bounding boxes.
Recent works distinguish between several settings based on the amount of annotations required.
The approaches either use part annotations, only bounding box annotations, or no annotation at all.
In addition, the required annotation in training is distinguished from the annotation required at test time.
Our approach only uses the class labels of the training images without additional annotation.

\myparagraph{CUB200-2001}
 \begin{table}
  \centering
  \resizebox{\linewidth}{!}{
    \begin{tabular}{cclc}
    \toprule
    Train. &Test  & Method & Accuracy\tabularnewline
    Anno.     &Anno.  &  & \tabularnewline
    \midrule
    Parts&Bbox&Bbox CNN features& 56.00\%\tabularnewline
    Parts&Bbox&Berg \etal~\cite{berg2013partclass}& 56.78\%\tabularnewline
    Parts&Bbox&Goering \etal~\cite{Goering14:NPT}& 57.84\%\tabularnewline
    Parts&Bbox&Chai \etal~\cite{chai2013symbiotic}& 59.40\%\tabularnewline
    Parts&Bbox&Simon~\etal~\cite{Simon14:PDD} & 62.53\%\tabularnewline
    Parts&Bbox&Donahue \etal~\cite{donahue2013decaf}& 64.96\%\tabularnewline
    \midrule
    Parts& None&Simon~\etal~\cite{Simon14:PDD} & 60.55\%\tabularnewline
    Parts& None&Zhang~\etal~\cite{zhang2014partRCNN} & 73.50\%\tabularnewline
    Parts& None&Branson~\etal~\cite{branson14cub75acc} & 75.70\%\tabularnewline
    \midrule
    Bbox& None&Simon~\etal~\cite{Simon14:PDD} & 53.75\%\tabularnewline
    \midrule
    None& None&Xaio~\etal~\cite{xiao14attention} (AlexNet)& 69.70\%\tabularnewline
    \vspace{7pt}None& None&Xaio~\etal~\cite{xiao14attention} (VGG19)& 77.90\%\tabularnewline
    None& None&No parts (AlexNet)& 52.20\%\tabularnewline
    None& None&Ours, rand., Sect.~\ref{sec:random_selection} (AlexNet)& $60.30\pm 0.74\%$\tabularnewline
    None& None&Ours, const., Sect.~\ref{sec:part_constellation} (AlexNet)& 68.50\%\tabularnewline
    None& None&No parts (VGG19)& 71.94\%\tabularnewline
    None& None&Ours, rand., Sect.~\ref{sec:random_selection} (VGG19)& $79.44\pm 0.56\%$\tabularnewline
    None& None&Ours, const., Sect.~\ref{sec:part_constellation} (VGG19)& \textbf{81.01\%}\tabularnewline
    \bottomrule
    \end{tabular}
   }
    \caption{Species categorization performance on CUB200-2011. 
    \label{tab:cub200_results}}
  \end{table}
The results of fine-grained recognition on CUB200-2011 are shown in Tab.~\ref{tab:cub200_results}.
We present three different results for every CNN architecture. 
``No parts'' corresponds to global image features only. 
``Ours, rand.'' and ``Ours, const.'' are the approaches presented in Sect.~\ref{sec:random_selection} and \ref{sec:part_constellation}.
As can be seen in the table, our approach improves the work of Xiao~\etal~\cite{xiao14attention} by 3.1\%, an error decrease of more than 16\%. 
It is important to note that their work requires a pre-trained classifier for birds in order to select useful patches for fine-tuning. 
In addition, the authors confirmed that they used a much larger bird subset of ImageNet for pre-training of their CNN.
In contrast, our work is easier to adapt to other datasets as we only require a generic pre-trained CNN and no domain specific outside training data.
The gap between our approach and the third best result in this setting by Simon~\etal~\cite{Simon14:PDD} is even higher with more than 27\% difference.
The table also shows results for the use of no parts and random part selection.
As can be seen, even random part selection improves the accuracy by 8\% on average compared to the use of no parts.
The presented part selection scheme boosts the performance even further to 68.5\% using AlexNet and 81.01\% using VGG19.

\myparagraph{NA birds}
 \begin{table}
  \centering
  \resizebox{\linewidth}{!}{
    \begin{tabular}{cclc}
    \toprule
    Train. &Test  & Method & Accuracy\tabularnewline
    Anno.     &Anno.  &  & \tabularnewline
    \midrule
    Parts&Parts&Horn~\etal~\cite{horn15nabirds}& 75.0\%\tabularnewline
    \midrule
    None& None&No parts (GoogLeNet)& 63.9\%\tabularnewline
    None& None&Ours, const., Sect.~\ref{sec:part_constellation} (GoogLeNet)& \textbf{76.3\%}\tabularnewline
    \bottomrule
    \end{tabular}
   }
    \caption{Species categorization performance on NA Birds. 
    \label{tab:nabirds_results}}
  \end{table}
The results of our approach on the relatively new NA birds dataset are shown in Tab.~\ref{tab:nabirds_results}.
The accuracy without using any parts is only 63.9\%.
Similar to the CUB200-2011 dataset, there is a clear advantage of using parts selected by our approach with an accuracy of 76.3\%. 
Interestingly, the accuracy is very close to the one on CUB200, while there are more than 2.5 times more classes in NA birds. 
We outperform the baseline provided the authors using the approach of \cite{branson14cub75acc} even though we are not using any kind of part annotation.

\myparagraph{Stanford dogs}
  \begin{table}
  \centering
  \scalebox{0.885}{%
    \begin{tabular}{lc}
    \toprule
    Method & Accuracy\tabularnewline
    \midrule
    Chai~\etal~\cite{chai2013symbiotic}& 45.60\%\tabularnewline
    Gavves~\etal~\cite{gavves2013alignment}& 50.10\%\tabularnewline
    Chen~\etal~\cite{chen15selPooling}& 52.00\%\tabularnewline
    \vspace{7pt}Google LeNet ft~\cite{sermanet2014attention}& 75.00\%\tabularnewline
    No parts (AlexNet)& 55.90\%\tabularnewline
    Ours, rand., Sect.~\ref{sec:random_selection} (AlexNet)&  $63.29\pm 0.97\%$\tabularnewline
    Ours, const., Sect.~\ref{sec:part_constellation} (AlexNet)& 68.61\%\tabularnewline
    \bottomrule
    \end{tabular}
   }
    \caption{Species categorization performance on Stanford dogs.
    \label{tab:dogs_results}}
  \end{table}
The accuracy on Stanford dogs is given in Tab.~\ref{tab:dogs_results}.
To the best of our knowledge, there is only one work showing results for a CNN trained from scratch excluding the testing images of Stanford dogs.
Sermanent~\etal~\cite{sermanet2014attention} fine-tuned the architecture of their very deep Google LeNet to obtain 75\% accuracy.
In our experiments, we used the much weaker architecture of Krizhevsky~\etal and still reached 68.61\%.
Compared to the other non-deep architectures, this means an improvement of more than 16\%.

\myparagraph{Oxford pets and flowers}
  \begin{table}
  \centering
  \scalebox{\tablescalingfactor}{
    \begin{tabular}{lc}
    \toprule
    Method & Accuracy\tabularnewline
    \midrule
    Angelova~\etal~\cite{angelova13flower}& 80.66\%\tabularnewline
    Murray~\etal~\cite{murray14catsFlowers}& 84.60\%\tabularnewline
    Razavian~\etal~\cite{razavian14cnnFeatures}& 86.80\%\tabularnewline
    \vspace{7pt}Azizpour~\etal~\cite{azizpour14CNNanalysis}& 91.30\%\tabularnewline
    No parts (AlexNet)& 90.35\%\tabularnewline
    Ours, rand., Sect.~\ref{sec:random_selection} (AlexNet)& $90.32\pm 0.18\%$\tabularnewline
    Ours, const., Sect.~\ref{sec:part_constellation} (AlexNet)& 91.74\%\tabularnewline
    No parts (VGG19)& 93.07\%\tabularnewline
    Ours, rand., Sect.~\ref{sec:random_selection} (VGG19)& $94.20\pm 0.23\%$\tabularnewline
    Ours, const., Sect.~\ref{sec:part_constellation} (VGG19)& \textbf{95.34\%}\tabularnewline
    \bottomrule
    \end{tabular}
   }
    \caption{Classification performance on Oxford 102 flowers.
    \label{tab:flowers_results}}
  \end{table}
  \begin{table}
  \centering
  \scalebox{\tablescalingfactor}{
    \begin{tabular}{lc}
    \toprule
    Method & Accuracy\tabularnewline
    \midrule
    Bo~\etal~\cite{bo13cats}.& 53.40\%\tabularnewline
    Angelova~\etal~\cite{angelova13flower}.& 54.30\%\tabularnewline
    Murray~\etal~\cite{murray14catsFlowers}.& 56.80\%\tabularnewline
    \vspace{7pt}Azizpour~\etal~\cite{azizpour14CNNanalysis}.& 88.10\%\tabularnewline
    No parts (AlexNet)&78.55\%\tabularnewline
    Ours, rand., Sect.~\ref{sec:random_selection} (AlexNet)& $82.70\pm 1.64\%$\tabularnewline
    Ours, const., Sect.~\ref{sec:part_constellation} (AlexNet)& 85.20\%\tabularnewline
    No parts (VGG19)& 88.76\%\tabularnewline
    Ours, rand., Sect.~\ref{sec:random_selection} (VGG19)& $90.42\pm 0.94\%$\tabularnewline
    Ours, const., Sect.~\ref{sec:part_constellation} (VGG19)& \textbf{91.60\%}\tabularnewline
    \bottomrule
    \end{tabular}
   }
    \caption{Species categorization performance on Oxford-IIIT Pets.
    \label{tab:cats_dogs_results}}
  \end{table}
The results for the Oxford flowers and pets dataset are shown in Tab.~\ref{tab:flowers_results} and \ref{tab:cats_dogs_results}.
Our approach consistently outperforms previous work by a large margin on both datasets. 
Similar to the other datasets, randomly selected parts already improve the accuracy by up to 4\%. 
Our approach significantly improves this even further and achieves 95.35\% and 91.60\%, respectively.

\myparagraph{Influence of the number of parts}
  \begin{table}
  \small
  \centering
      \resizebox{0.9\linewidth}{!}{
	  \begin{tikzpicture}
	    \begin{axis}[
		    xlabel=Number of parts used,
		    ylabel style={align=center},ylabel=Accuracy in \%\\on CUB200-2011,
		    ymin = 69,
		    ymax = 82,
		    legend pos=south east,
		    width=\linewidth,
		    height=0.6\linewidth]
	    \addplot[color=red,mark=x] coordinates {
		    (0, 70.8)
		    (1, 76.9)
		    (2, 78.6)
		    (4, 79.0)
		    (5, 79.5)
		    (6, 79.2)
		    (9, 79.4)
		    (10, 79.6)
		    (16, 79.6)
		    (25, 79.8)
		    (36, 79.8)
		    (49, 79.8)
		    (64, 79.4)
		    (81, 79.5)
		    (100, 79.7)
		    (256, 79.168)
	    };

	    \addplot[color=blue,mark=+] coordinates {
		    (0, 70.8)
		    (1, 73.0296)
		    (2, 72.7764)
		    (3, 75.0547)
		    (4, 75.1870)
		    (5, 74.6116)
		    (6, 75.1237)
		    (9, 75.6587)
		    (10, 76.5907)
		    (22, 77.5285)
		    (36, 77.4019)
		    (60, 78.4950)
		    (80, 78.5899)
		    (100, 78.5468)
		    (256, 79.1681 )
	    };

	    \legend{{Ours, constellation},{Ours, random parts}}
	    \end{axis}
	  \end{tikzpicture}
      }
    \caption{Influence of the number of parts on the accuracy on CUB200-2011.
    One patch was extracted for each part proposal. 
    \label{tab:influence_parts}}
  \end{table}
Fig.~\ref{tab:influence_parts} provides insight into the influence of the number of parts used in classification.
We compare to random part to the part constellation model based selection.
In contrast to the previous experiments, one patch is extracted per part using $\lambda=\frac{1}{10}$.
While random parts increase the accuracy for any amount of parts, the presented scheme clearly selects more relevant parts and helps to greatly improve the accuracy.

    \subsection{From fine-grained to generic classification}
  \begin{table}
  \centering
  \scalebox{\tablescalingfactor}{
    \begin{tabular}{lc}
    \toprule
    Method & Accuracy\tabularnewline
    \midrule
    Zeiler~\etal~\cite{zeiler2013visualizing}& 74.20\%\tabularnewline
    Chatfield~\etal~\cite{chatfield14return}& 78.82\%\tabularnewline
    \vspace{7pt}Simonyan~\etal~\cite{Simonyan14veryDeep} + VGG19& $85.10\%$\tabularnewline
    No parts (AlexNet)& 71.44\%\tabularnewline
    Ours, rand., Sect.~\ref{sec:random_selection} (AlexNet)& $72.39\%$\tabularnewline
    Ours, const., Sect.~\ref{sec:part_constellation} (AlexNet)& $72.57\%$\tabularnewline
    No parts (VGG19)& $82.44\%$\tabularnewline%
    Ours, const., Sect.~\ref{sec:part_constellation} (VGG19)& $84.10\%$\tabularnewline
    \bottomrule
    \end{tabular}
   }
    \caption{Accuracy on the Caltech 256 dataset with 60 training images per category. 
    \label{tab:cal256_res}}
  \end{table}

Almost all current approaches in fine-grained recognition are specialized algorithms and it is hardly possible to apply them to generic classification tasks. 
The main reason is the common assumption in fine-grained recognition that there are shared semantic parts for all objects. 
Does that mean that all the rich knowledge in the area of fine-grained recognition will never be useful for other areas?
Are fine-grained and generic classification so different?
In our opinion, the answer is a clear no and the proposed approach is a good example for that.

There are two main challenges for applying fine-grained classification approaches to other tasks.
First, the semantic part detectors need to be replaced by more abstract interest point detectors.
Second, the selection or training of useful interest point detectors needs to consider that each object class has its own unique shape and set of semantic parts.
Our approach can be applied to generic classification tasks in a natural way.
The first challenge is already solved by using the part detectors of a CNN trained to distinguish a huge number of classes.
Because of these properties, part proposals can be seen as generic interest point detectors with a focus on a special pattern.
In contrast to semantic parts, they are not necessarily only recognizing a specific part of a specific object. 
Instead, they capture interesting points of many different kinds of objects.
The second challenge is tackled by building class-wise part models and selecting part proposals that are shared among most classes.
However, even a random selection of part detectors turns out to increase the classification accuracy already.

\myparagraph{Caltech 256}
The results of our approach on Caltech 256 are shown in Tab.~\ref{tab:cal256_res}.
The proposed methods improves the baseline of global features without oversampling by $1\%$ in case of AlexNet and 1.6\% in case of VGG19. 
While Simonyan~\etal achieves slightly higher performance, their approach is also much more expensive due to dense evaluation of the whole CNN over all possible crops at three different scales.
Their best result of $86.2\%$ is achieved by using a fusion of two CNN models, which is not done in our case and consequently not comparable.
The results clearly shows that replacing semantic part detectors by more generic detectors can be enough to apply fine-grained classification approaches in other areas. 
Many current approaches in generic image classification rely on ``blind'' parts.
For example, spatial pyramids or other oversampling methods are equivalent to part detectors that always detect something at a fixed position in the image. 
Replacing these ``blind'' detections by more sophisticated ones 
in combination with class-wise part models is a natural improvement.

\section{Conclusions\label{sec:conclusions}}
\vspace{-5pt}
This paper presents an unsupervised approach for the selection of generic parts for fine-grained and generic image classification.
Given a CNN pre-trained for classification, we exploit the learned inherit part detectors for generic part detection.
A part constellation model is estimated by analyzing the predicted part locations for all training images. 
The resulting model contains a selection of useful part proposals as well as their spatial relationship in different views of the object of interest. 

We use this part model for part-based image classification in fine-grained and generic object recognition.
In contrast to many recent fine-grained works, our approach surpasses the state-of-the-art in this area and is beneficial for other tasks like data augmentation and generic object classification as well. 
This is supported by, among other results, a recognition rate of 81.0\% on CUB200-2011 without additional annotation and $84.1\%$ accuracy on Caltech 256.

In our future work, we plan to use the deep neural activation maps directly as probability maps while maintaining the speed of our current approach.
The estimation of object scale would allow for applying our approach to datasets in which objects only cover a small part of the image.
Our current limitation is the assumption that a single channel corresponds to a object part. 
A combination of channels can be considered to improve localization accuracy.
In addition, we plan to learn the constellation models and the subsequent classification jointly in a common framework.

\section{Changelog}
\begin{itemize}
 \item V3: Added results for NA birds
 \item V2: Updated to camera ready version
\end{itemize}

{\small
\bibliographystyle{ieee}
\bibliography{paper}
}

\end{document}